\documentclass[pra,reprint,superscriptaddress,amsmath,amssymb]{revtex4-2}
\usepackage{graphicx,booktabs,multirow,amsthm}
\usepackage[colorlinks=true,linkcolor=blue!50!black,citecolor=blue!50!black,urlcolor=blue!50!black]{hyperref}
\usepackage{xcolor}

\newtheorem{theorem}{Theorem}
\newcommand{\conj}[1]{\overline{#1}}
\newcommand{\PW}{\mathrm{PW}}

\newcommand{\NBeads}{53}

\newcommand{\FrameCxWins}{31}
\newcommand{\FrameCxTies}{20}
\newcommand{\FrameCxLosses}{2}
\newcommand{\FrameCxMed}{+0.0020}

\newcommand{\FrameReLosses}{28}
\newcommand{\FrameReMed}{-0.0021}

\newcommand{\FramePhotons}{164}

\newcommand{\FrameBgShare}{7.3}
\newcommand{\FrameBgDTest}{+4.6\times10^{-5}}
\newcommand{\FrameFcLo}{0.8}
\newcommand{\FrameFcHi}{1.35}
\newcommand{\FrameFcBest}{0.9}
\newcommand{\FrameFcGap}{4\times10^{-4}}

\newcommand{\FrameCert}{1.26}
\newcommand{\FrameStreamFc}{0.84}

\newcommand{\WinBgWins}{24}

\newcommand{\WinBgLosses}{9}
\newcommand{\WinBgN}{50}
\newcommand{\WinBgMed}{+0.0011}

\newcommand{\PhN}{12{,}767}

\newcommand{\ElN}{7{,}153}

\newcommand{\MmN}{223{,}514}

\newcommand{\GhN}{10{,}627}

\newcommand{\ElBetaMeas}{0.35}

\newcommand{\IndPhBound}{0.04}

\begin{document}

\title{Complex band-limited maximum likelihood: estimating the wave behind detected quanta}

\author{Rahul Agarwal}
\email{rahul.jhu@gmail.com}
\affiliation{Independent researcher, Sammamish, WA, USA} 
\author{Ankit Agarwal}
\email{ankit.iiith@gmail.com}
\affiliation{Independent researcher, San Jose, CA, USA}

\begin{abstract}
Detected photons and electrons land at positions drawn from $|\psi|^2$, yet it is $\psi$, phase included, that a
wave equation propagates, and the state that produced the particles is often known only in part. A density estimator can
recover $|\psi|^2$ directly from the positions, without any wave equation, but it discards the phase, and with it the
state. We introduce complex band-limited maximum likelihood
(complex BLML), which estimates $\psi$ itself by maximising the likelihood of the detected positions over exactly the intensities the physics
allows: squared moduli of complex functions whose spectrum lies in a given set, such as a band, the shape of the
aperture, or several modes for a partially coherent source. A single eigenvalue can certify a global
maximum. On simulated two-slit far fields we measure what each prior is worth: a band
reaches a given accuracy with seven times fewer photons than a kernel density estimate and already gives the fringe
contrast and the relative phase, and the shape of the aperture adds the slit geometry. Some quantities need a prior
at any number of photons: for slits of one profile the intensity fixes only the contrast $2\rho V/(1+\rho^2)$, and
on a line two different apertures can produce the same particles. Cutting a pattern to a few fringes, as published
records are, removes the slit width. On four public single-particle double-slit records it predicts held-out
particles best among the methods we tested, kernel estimators and two-slit optical models included, and it recovers the
fringe contrast without a model of the optics. On fluorescent beads it finds the resolution limit of the microscope,
$2\mathrm{NA}/\lambda$, from the photons alone.
\end{abstract}

\maketitle

\section{Introduction}
\label{sec:intro}

Quantum mechanics takes the wavefunction as its basic object, and its dynamics, the Schr\"odinger equation or, for
light, the wave equation of the field, act on $\psi$ with its phase. Many experiments, however, record only where
particles land. A detector that counts single quanta, such as photons on an intensified camera or electrons on a
microchannel plate, records the position of each one, and by the Born rule each position is a random draw from
$|\psi|^2$. When the state is fully known, a wave equation predicts $|\psi|^2$ and can be tested against the positions
directly. Usually it is not. Even for a double slit, whose geometry is nominally known, the geometry fixes only where the
field is, not the field itself: the amplitude and phase of the illumination on each slit, the coherence of the source,
the true slit profile and any defocus are known only in part, and a camera records none of the phase. What the particles then say about the wave is $\psi$ itself, the state the equation propagates, and recovering it
with its phase, which one intensity does not show, is what an estimate from the detected positions must do. With
records in two planes, such an estimate also tests the propagation itself. The wavefunction can be reached more directly, by weak measurements
\cite{lundeen2011direct} or by interference with a reference wave, but these need additional optics; records of
positions are the simplest and most common data. How far they determine $\psi$ is an old question: Pauli asked whether
the position and momentum distributions together fix the wavefunction, and in general they do not
\cite{corbett1978wave}.

What makes the problem tractable is that the Fourier transform of $\psi$, the wavefunction in the conjugate domain,
usually has a limited support. On its way to the detector the wave passes an aperture, for example a pair of slits or
the pupil of a microscope. In the far field the wave is the Fourier transform of that aperture, so it is
band-limited, with a band set by the size of the aperture \cite{goodman2005fourier,born1999principles}. The density
of detected particles is therefore the squared modulus of a band-limited complex function. For a partially coherent
source it is a sum of several such terms \cite{wolf1982new}. This is the only assumption the estimator below needs,
and few density estimation problems come with such a specific one.

Records of this kind are usually analysed in one of three ways, and each uses only part of what the physics provides.
A histogram or a kernel density estimate (KDE) uses none of it. A parametric model of the optics uses all of it, but
it needs the wavelength, the distances and the exact profile of the aperture, and when one of these is wrong the fit
is biased without showing it. Phase retrieval \cite{gerchberg1972practical,fienup1982phase,gonsalves1982phase}
recovers the field from the intensity when the support of the aperture is given, and can do so by maximum Poisson
likelihood even at very low counts \cite{li2022poisson}. It does not, however, return a density that can be tested on
held-out data, and it offers no rule for choosing the support. Partial coherence needs yet other tools, such as
mixed-state ptychography \cite{thibault2013reconstructing}, which needs many overlapping exposures, or single-shot
coherence measurement \cite{albrecht2023single}, which needs a known mask and a bright source.

We introduce an estimator that maximises the likelihood over exactly the intensities that the physics allows. The
model is $f=\sum_k|g_k|^2$, where each $g_k$ is a complex function whose spectrum lies in a set given by the physics.
We call it complex BLML: it extends band-limited maximum likelihood (BLML) \cite{agarwal2017blml}, which fits a real
root, to complex roots and to several modes with their own kernels. It relates to a continuous field
the way maximum-likelihood state tomography \cite{hradil1997quantum} relates to a finite-dimensional quantum state.
The weakest useful prior is a band. A stronger one is the shape of the aperture: in the far field the spectrum of
$\psi$ is the aperture itself, so for two slits it lies on the two slits, and for a microscope on the pupil disk. A
mode with a much narrower band than the aperture describes a smooth background. The kernel, the number of modes and
the band are all chosen by held-out likelihood. Beyond the shape of the opening, no model of the optics is needed: not
the profile of the slits, not their phase, and not their amplitudes.

A single fit returns several things. It is a density estimate, and on the data in this paper it is the most accurate
of those we compare. Its root estimates the field in the aperture, including its phase, which no density estimator
provides. The fitted kernel gives the geometry of the aperture, here the separation and width of the slits. With
several modes that share one geometry, it splits the light into a coherent and an incoherent part. In this way one
likelihood can carry priors of very different strength: smoothness, a band, the shape of the aperture, its exact
geometry, or a coherence structure. Tasks that are normally done by separate methods, such as kernel estimation,
band-limited density estimation, phase retrieval with a support and coherent-mode decomposition, become the same fit
with a different kernel (Table~\ref{tab:rungs}).

This lets an experimenter ask two practical questions before fitting. Which unknowns does the intensity determine
under a given prior? And how many particles does each prior save? We answer both for the double slit. We also answer
a third question that matters for published records: what is lost when the pattern is cut down to a few fringes?

Section~\ref{sec:model} describes the estimator and what one intensity can determine: on a line the band alone does
not fix which complex root is physical, for two slits of one profile only the fringe contrast
$\gamma=2\rho V/(1+\rho^2)$ is fixed, and a window of $L$ pixels blurs the aperture over $1/L$. Section~\ref{sec:synth}
measures what each prior is worth on a known truth, recorded whole and cut to a few fringes. Section~\ref{sec:dslit}
applies the estimator to four published single-particle double-slit records, photons from a coherent, a
ghost-diffraction and a partially coherent source \cite{aspden2016video} and electrons \cite{bach2013controlled}, with
replicas of each record as a calibrated reference, and Sec.~\ref{sec:beads} to a microscope imaging fluorescent beads.

A band limit at the detector is a support constraint on the aperture, so complex BLML is also a form of phase
retrieval: by maximum likelihood on the detected particles, with the support chosen from held-out data. Maximum
likelihood has been used to refine phase-retrieval reconstructions of coherent diffraction data
\cite{thibault2012maximum,li2022poisson}, and the non-convex fit of a field to intensities has been studied as
Wirtinger flow \cite{candes2015wirtinger}, with a convex relaxation over positive semidefinite matrices, PhaseLift
\cite{candes2013phaselift}. Our certificate (Sec.~\ref{sec:model}) is the optimality condition of a lift of the same
kind, applied to the likelihood of the detected particles with a kernel that encodes the prior; it certifies the
non-convex fit instead of replacing it. Sums of squares
of functions in a reproducing kernel Hilbert space are the positive semidefinite models of
\cite{marteauferey2020nonparametric}; here the kernel comes from the physics and each mode has a physical meaning.

\section{The estimator}
\label{sec:model}

\subsection{Band-limited fields and their modes}

Let the aperture $\Omega\subset\mathbb R^d$ be a compact, symmetric set: an interval $[-f_c/2,f_c/2]$ on the line
(the band), a pair of slits, or a disk $|\nu|\le f_c/2$ for a circular pupil. The functions whose spectrum lies in
$\Omega$ form the Paley--Wiener space
\[
\PW_\Omega=\bigl\{g\in L^2(\mathbb R^d;\mathbb C):\ \hat g(\nu)=0\text{ for }\nu\notin\Omega\bigr\},
\]
where $\hat g(\nu)=\int g(x)e^{-2\pi i\nu\cdot x}\,dx$. This space has the reproducing kernel
$K(x)=\int_\Omega e^{2\pi i\nu\cdot x}\,d\nu$, which is real and even and satisfies $g(y)=\langle g,K(\cdot-y)\rangle$.
On the line the kernel is $K(x)=f_c\,\mathrm{sinc}(f_cx)$. For the disk it is the jinc kernel, and for a periodic
coordinate it is the normalised Dirichlet kernel with a whole number of harmonics. For two equal slits of width $a$ at
separation $d$, the aperture is $\Omega=\{\nu:\ |\,|\nu|-\nu_d\,|\le\nu_a/2\}$, and the kernel is the difference of
two sinc kernels,
\begin{equation}
K(t)=f_c\,\mathrm{sinc}(f_ct)-f_i\,\mathrm{sinc}(f_it),\qquad f_i=(1-w)f_c.
\label{eq:slitkernel}
\end{equation}
Here $f_c$ is the full width of the aperture, and the fill $w=2a/(d+a)$ is the fraction of it that is open; $w=1$
gives back the band. This kernel assumes that there are two slits, but nothing about their profile, phase or
amplitudes. Like the band, $f_c$ and $w$ are chosen by held-out likelihood. When a record is cut to a window of $L$
pixels and the window is treated as one period, the same aperture becomes a set of harmonics $k/L$ with
$k_{\rm lo}\le|k|\le k_{\rm hi}$. The periodic two-slit kernel is then the difference of two Dirichlet kernels, and it
is normalised on the window instead of on the whole line (Sec.~\ref{sec:crop}). In the same way, a pupil on a window that is
periodic in both axes is the disk of harmonics $\|k/P\|\le f_c/2$, whose kernel is the ball kernel made periodic.

The model is
\begin{equation}
f=\sum_{k=1}^K|g_k|^2,\qquad g_k\in\PW_{\Omega_k},\qquad\sum_k\|g_k\|^2=1 .
\label{eq:model}
\end{equation}
By Plancherel's theorem $\int f=\sum_k\|g_k\|^2$, so the constraint makes $f$ a probability density, and the spectrum
of $f$ lies in $\bigcup_k(\Omega_k-\Omega_k)$. Real BLML is the case of one mode with $g$ real. Complex BLML is one mode
with $g$ complex, which is the same as two real modes, $g=a+ib$. Equation~\eqref{eq:model} is also how intensities
arise in optics. A partially coherent field has a cross-spectral density
$W(x,y)=\sum_k\lambda_k\phi_k(x)\conj{\phi_k(y)}$ in which each coherent mode is band-limited by the pupil, and the
intensity is its diagonal \cite{wolf1982new}. Because each mode can have its own band, one mode can be sharp and
another broad. We use this for the background: a mode with a periodic kernel of $N$ harmonics across the record is a
smooth function with at most $N$ cycles over it ($N=0$ is a constant), and its weight $\|g_{\rm bg}\|^2$ is the
background fraction, fitted together with everything else.

The same likelihood can therefore carry priors of very different strength, and the prior enters only through the
kernel of each mode and the number of modes. Table~\ref{tab:rungs} lists the priors used in this paper. For each one
it gives the kernel that expresses it, the method that is normally used for that kind of knowledge, and what the fit
returns.

\begin{table*}[t]
\centering\small
\caption{One likelihood, many priors. Complex BLML maximises the same likelihood, Eq.~\eqref{eq:model}, whatever is
known about the aperture; only the kernel and the number of modes change. Rows 2--7 are this one estimator. Row 1, a
kernel density estimate, is the baseline that uses no physics, and row 8, a parametric model of the optics, is the
limit where everything is known. The last column names the version used in Sec.~\ref{sec:synth} (the curves of
Figs.~\ref{fig:ladder} and \ref{fig:crop}) and in Secs.~\ref{sec:dslit} and \ref{sec:beads}.}
\label{tab:rungs}
\begin{tabular}{@{}p{0.03\linewidth}p{0.16\linewidth}p{0.2\linewidth}p{0.2\linewidth}p{0.22\linewidth}p{0.13\linewidth}@{}}
\toprule
& what is known & kernel / modes & usually handled by & what the fit returns & used here as \\
\midrule
1 & smoothness only & -- & kernel density estimate, histogram & the intensity & KDE \\[3pt]
2 & the extent of the aperture & one band: sinc (line), jinc (disk), Dirichlet (window) & band-limited density estimation (real BLML \cite{agarwal2017blml}) & the intensity, and a field that is not unique on the line & band; beads at $2\mathrm{NA}/\lambda$ \\[3pt]
3 & the shape of the aperture (its support) & the kernel of that support, e.g.\ two slits, Eq.~\eqref{eq:slitkernel}, or a disk & phase retrieval with a support \cite{fienup1982phase} & also the field in the aperture: geometry, relative phase, amplitudes & two-slit shape \\[3pt]
4 & its exact geometry & the same kernel, with fixed parameters & phase retrieval with a known support & the field in the aperture & known geometry \\[3pt]
5 & partial coherence & several modes sharing the aperture & coherent-mode and mixed-state methods \cite{wolf1982new,thibault2013reconstructing} & also a split into coherent and incoherent light (only $\gamma$ without a further prior) & three modes; multimode record \\[3pt]
6 & a smooth background & one extra mode with a few harmonics & background subtraction, a fitted floor & the background fraction & all fits \\[3pt]
7 & a finite record & the kernel made periodic on the window & -- & as above, on the window & periodic kernel; the records \\[3pt]
8 & the full optics & a parametric model (not BLML) & Fraunhofer and Fresnel fits, PSF models & a few parameters, if the model is right & full physical model \\
\bottomrule
\end{tabular}
\end{table*}

\subsection{Stationarity and the coordinate update}

Given samples $x_1,\dots,x_n$ from $f$, we maximise the log-likelihood $\ell=\sum_i\log f(x_i)$ under the constraint in
\eqref{eq:model}. Projecting each $g_k$ onto the span of its kernel at the samples keeps its values at the samples and
does not increase its norm. Every maximiser is therefore a finite expansion $g_k=\frac1n\sum_ic_{ik}K_k(\cdot-x_i)$.
Let $S_k$ be the Gram matrix of $K_k$ at the samples, let $C\in\mathbb C^{n\times K}$ hold the coefficients, with
columns $c_k$ and rows $C_i$, and let $\|C_i\|^2=\sum_k|c_{ik}|^2$.

\begin{theorem}[Stationarity]
\label{thm:stat}
At every critical point (after replacing $C$ by coefficients that define the same modes if some $S_k$ is singular),
\begin{equation}
(S_kc_k)_i=n\,\frac{c_{ik}}{\|C_i\|^2}\qquad\text{for every sample }i\text{ and mode }k .
\label{eq:stat}
\end{equation}
Then $f(x_i)=1/\|C_i\|^2$, the constraint holds automatically, and $\ell=-\sum_i\log\|C_i\|^2$. For one mode,
\eqref{eq:stat} reads $c_i\,\conj{(Sc)_i}=n$, so $\arg c_i=\arg g(x_i)$ and $|c_i|\,|g(x_i)|=1$; for real $c$ it is
the BLML system $c_i(Sc)_i=n$.
\end{theorem}

The proof (Supplemental Material, Sec.~S1) uses Wirtinger derivatives. Every real sign pattern also gives a solution
of \eqref{eq:stat}, so the complex maximum is at least as high as the best sign pattern: a sign is a phase of $0$ or
$\pi$, and the phases of complex BLML can move continuously between sign patterns. We solve \eqref{eq:stat} by a
coordinate update, which generalises that of real BLML \cite{agarwal2026highdim}, or by L-BFGS on the coefficients,
on CPU and GPU, starting from the real positive solution with small random phases (Supplemental Material, Sec.~S2).

\subsection{A certificate of global optimality}

When all modes share one kernel, every number of modes is a special case of one convex problem. Write
$f(x)=\kappa(x)^\top A\,\kappa(x)$ with $\kappa(x)=(K(x-x_j))_j/n$ and $A\succeq0$ a real symmetric matrix. Real BLML
is $A=aa^\top$, complex BLML is $A=aa^\top+bb^\top$, and $K$ complex modes give $A=\mathrm{Re}(CC^H)$. With $s_i$ the
$i$-th column of $S$, maximum likelihood over all $A$ is $\max_{A\succeq0}\sum_i\log(s_i^\top As_i)$ subject to
$\mathrm{tr}(AS)=n^2$: a concave objective with a linear constraint.

\begin{theorem}[Certificate]
\label{thm:cert}
Let $C$ satisfy \eqref{eq:stat} with a shared kernel and $D=\mathrm{diag}(\|C_i\|^2)$. If
\begin{equation}
\lambda_{\max}\bigl(D^{1/2}SD^{1/2}\bigr)\le n ,
\label{eq:cert}
\end{equation}
then $A=\mathrm{Re}(CC^H)$ is a global maximum: no sign pattern, no complex root and no sum of squares attains a
larger likelihood.
\end{theorem}

The proof checks the Karush--Kuhn--Tucker conditions (Supplemental Material, Sec.~S1). The certificate costs one top
eigenvalue, and it is also the second-order test of the complex problem at a real solution, which is how the complex
iteration escapes a wrong sign pattern (Supplemental Material, Sec.~S2).

\subsection{What one intensity determines}

On the line, a band-limited density always has a band-limited complex root, but it rarely has a real one.

\begin{theorem}[Square roots on the line]
\label{thm:realroot}
Let $f\ge0$ be integrable on $\mathbb R$ with spectrum in $[-f_c,f_c]$, and $F$ its entire extension. Then $f=|g|^2$
for some $g\in\PW_{[-f_c/2,f_c/2]}$, and $f=g^2$ for a real such $g$ if and only if every zero of $F$ in $\mathbb C$
has even multiplicity.
\end{theorem}

The first part is Akhiezer's factorisation $F=GG^*$ \cite{akhiezer1956theory,boas1954entire}; the second is proved in
the Supplemental Material, Sec.~S1. Real zeros of $F$ are automatically double, because $f\ge0$. The condition
therefore concerns the non-real zeros, which come in conjugate pairs, and a small perturbation of $f$ generically makes
them simple. The in-focus slit, whose amplitude is real, is an exception; a defocused slit is not. Complex BLML is
therefore correctly specified for every band-limited density on the line, coherent or not, while the positive root of
real BLML is not. For an in-focus slit, the positive root has a corner at every dark fringe and 8\% of its spectral
energy outside the physical band (Supplemental Material, Sec.~S3).

The root is not unique. If $g$ is a maximiser, so are $e^{i\theta}g$ and its twin $\conj g$, whose transform is
$\conj{\hat g(-\nu)}$. On the line, so is every root obtained by moving one zero of $G$ to its complex conjugate
\cite{walther1963question,hofstetter1964construction}. So is $g\,e^{2\pi isx}$, which shifts the aperture by $s$: the
intensity fixes the geometry of the aperture but not its position, and only a band that fits the aperture closely
leaves little room for this shift. The twin flips the sign of the even aberrations, so one intensity image cannot give it; a second
image at a known defocus can \cite{gonsalves1982phase}. In two or more dimensions the global phase, the twin and a
translation are generically the only ambiguities \cite{bruck1979ambiguity,hayes1982reconstruction}, but a single root
need not exist, which is a second reason to allow several modes (Supplemental Material, Sec.~S1). An aperture smaller than the band removes most of the alternative roots on the line, and an
aperture made of well-separated parts, such as two slits, makes the root unique up to the trivial ambiguities
\cite{crimmins1983uniqueness}: moving a zero of $G$ to its conjugate changes the spectrum of the root, which then
leaves the aperture (Sec.~\ref{sec:dbench}).

Two further limits are specific to the double slit, and both are exact. The first concerns amplitudes and coherence.
If the two slits have the same profile, their far fields have the same modulus, $|u_2|=|u_1|$, and a beam with
amplitude ratio $\rho$, relative phase $\phi$ and visibility $V$ produces the intensity
\begin{equation}
f(x)\propto|u_1(x)|^2\bigl[1+\rho^2+2\rho V\cos(2\pi dx+\phi)\bigr].
\label{eq:gamma}
\end{equation}
This depends on $\rho$ and $V$ only through the fringe contrast $\gamma=2\rho V/(1+\rho^2)$. Coherent light on unequal
slits and partially coherent light on equal slits therefore produce exactly the same particles. Only a further prior
separates them: a coherent source ($V=1$) gives $\rho$, and equal slits ($\rho=1$) give $V$. Defocus breaks this
symmetry, because it moves the envelopes of the two slits apart, but only weakly: with $0.8$\,rad of quadratic phase
at the slit centres, the two descriptions differ by at most 9\% of the peak intensity.

The second limit is the window. A detector, or an analyst, keeps only a window of $L$ pixels. The data are then drawn
from the windowed intensity, whose root is $g\,\mathbb 1_L$, and the spectrum of that root is
$A*L\,\mathrm{sinc}(L\nu)$: the aperture blurred over a scale of $1/L$. A slit narrower than a few multiples of $1/L$
is then no longer resolved, although the slit separation, the fringe contrast and the fringe phase still are. A kernel
on the whole line causes a further problem: it normalises $|g|^2$ over the whole line, while the data lie only in the
window. On a window of a few fringes the likelihood then favours fields with less mass outside the window, which means
wider and smoother slits (Sec.~\ref{sec:crop}). A kernel that is periodic on the window has no outside. If the window
spans an odd number of half fringe periods, with both edges on dark fringes, the windowed field is close to periodic,
and its harmonics $k/L$ fall into two groups, one for each slit.

\section{Simulations}
\label{sec:synth}

We first test the estimator where the truth is known. We use one truth throughout and ask two questions. How many
particles does each prior save when the whole pattern is recorded? And what is lost when the pattern is cut to a few
fringes, as in the records of Sec.~\ref{sec:dslit}?

\begin{figure*}[t]
\centering
\includegraphics[width=\linewidth]{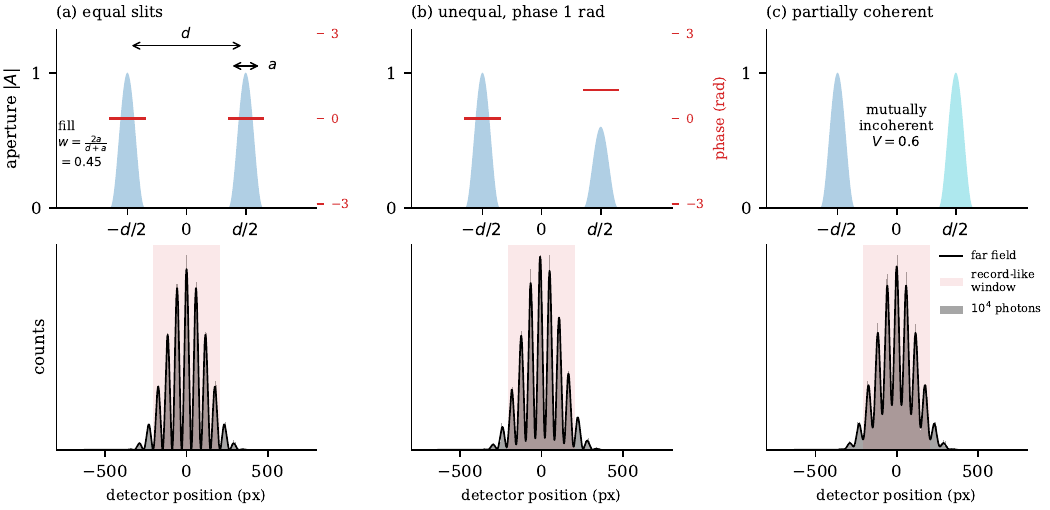}
\caption{The synthetic truths. Top: the aperture $A(\nu)$, which is the spectrum of the field, in units of the slit
separation $d$ (blue: $|A|$; red: its phase). There are two slits of width $a$, each tapered as $\cos^2$, with fill
$w=2a/(d+a)=0.45$. (a)~Equal slits. (b)~Unequal slits, with amplitude ratio $0.6$ and a phase step of $1$\,rad.
(c)~Equal slits that are mutually incoherent, with visibility $V=0.6$; this beam has no single field. Bottom: the far
field on a line of 2048 pixels, with a histogram of $10^4$ photons drawn from it. The shaded band is the record-like
window of seven fringes used in Sec.~\ref{sec:crop}.}
\label{fig:synthtruth}
\end{figure*}

\subsection{The prior ladder on an uncropped pattern}
\label{sec:dbench}

\paragraph{Truth and estimators.} We draw photons from the far field of two slits on a line of 2048 pixels
(Fig.~\ref{fig:synthtruth}). The slit centres are at $\pm0.0085$ cycles per pixel and the slit width is $0.005$. There
are three truths: equal slits, unequal slits with a phase step, and equal slits at visibility $0.6$. Each is simulated
without background and with a flat background of 10\%, at $n=10^3$ to $10^6$ photons, with three data sets per
setting. Because the slits are tapered, the field leaves only $4\times10^{-6}$ of its mass outside the line, so
nothing that $10^6$ photons could resolve is cut off.

The estimators form a ladder of increasingly strong priors, rows 1--5 and 8 of Table~\ref{tab:rungs}. Cross-validation
chooses the cut-off of the band, and the cut-off and fill of the two-slit kernel. The full physical model fits seven
numbers (the pattern centre, $d$, $a$, $\rho$, $\phi$, $V$ and the background) by maximum likelihood, started at the
truth. All complex BLML fits carry a flat background mode, and cross-validation is five-fold with the
one-standard-error rule. We score the intensity by its KL divergence from the truth. From the aperture $\hat A$, which
is the Fourier transform of the fitted root, we score the slit separation, the slit width, the relative phase of the
slits and their amplitude ratio (Supplemental Material, Sec.~S7). A linear phase on $g$ shifts the aperture as a whole
without changing $|g|^2$, so the position of the aperture is not determined, only its geometry; all scores are
insensitive to this shift. The fringe contrast $\gamma$ is scored on the density itself.

\begin{figure*}[t]
\centering
\includegraphics[width=\linewidth]{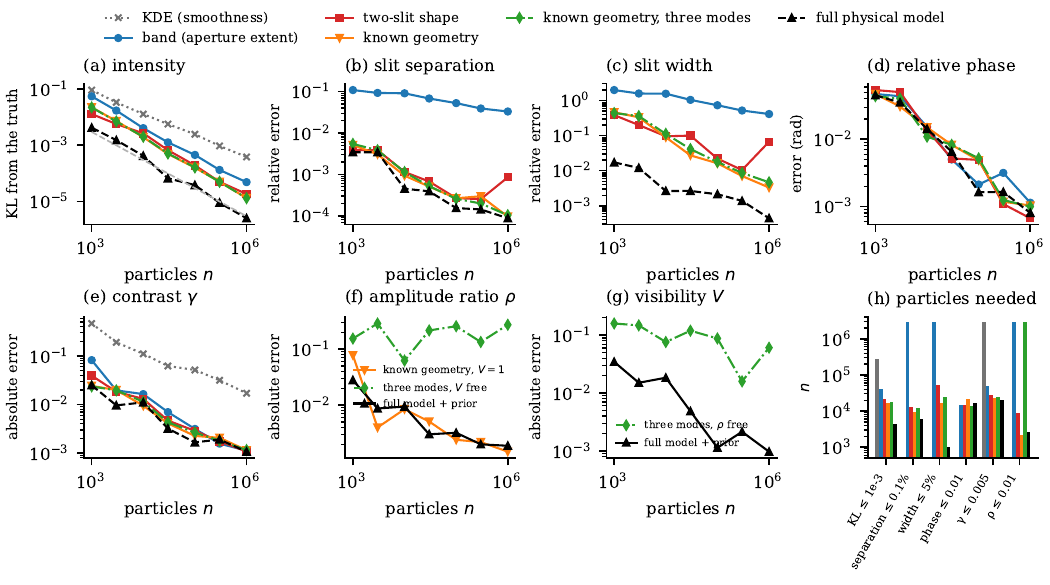}
\caption{The prior ladder on the uncropped pattern. Each panel shows an error against the number of photons, as the
median over six data sets per point (three without and three with a 10\% background), averaged geometrically over the
truths that the panel uses. (a)~KL divergence from the truth; the grey line has slope $1/n$. (b)~Slit separation and
(c)~slit width, as relative errors (the width for equal and unequal slits). (d)~Relative phase of the slits (unequal
slits). (e)~Fringe contrast $\gamma$ (unequal slits and the partially coherent truth). (f)~Amplitude ratio of the
unequal slits and (g)~visibility of the partially coherent truth. Each is shown with the prior that identifies it
($V=1$ or $\rho=1$) and from the three-mode model, which leaves the other quantity free; its error does not fall with
$n$. (h)~Photons needed to reach the targets of Table~\ref{tab:laddern}; bars at the top are not reached by $10^6$.}
\label{fig:ladder}
\end{figure*}

\begin{table}[t]
\centering\small
\caption{Photons each prior needs. Each entry is the number of photons $n$ at which the median error in
Fig.~\ref{fig:ladder} first falls below the target in the column head. For KL the target is a divergence from the true
density; for the others it is the absolute or relative error of the quantity. Values are interpolated on a log scale
between the simulated $n$, the smallest of which is $10^3$. The KDE returns no field. The full model is fitted with
$V=1$ for $\rho$; the three-mode model leaves $V$ free and does not determine $\rho$.}
\label{tab:laddern}
\resizebox{\linewidth}{!}{\begin{tabular}{lrrrrrr}
\toprule
prior & KL $\le10^{-3}$ & separation err.\ $\le0.1\%$ & width err.\ $\le5\%$ & phase err.\ $\le0.01$ & $\gamma$ err.\ $\le0.005$ & $\rho$ err.\ $\le0.01$ \\
\midrule
KDE (smoothness) & $2.8\times10^5$ & -- & -- & -- & $>10^6$ & -- \\
band (aperture extent) & $4.0\times10^4$ & $>10^6$ & $>10^6$ & $1.5\times10^4$ & $5.0\times10^4$ & $>10^6$ \\
two-slit shape & $2.2\times10^4$ & $1.3\times10^4$ & $5.2\times10^4$ & $1.4\times10^4$ & $2.8\times10^4$ & $8.7\times10^3$ \\
known geometry & $1.7\times10^4$ & $9.3\times10^3$ & $1.7\times10^4$ & $2.1\times10^4$ & $2.3\times10^4$ & $2.1\times10^3$ \\
known geometry, three modes & $1.8\times10^4$ & $1.2\times10^4$ & $2.4\times10^4$ & $1.4\times10^4$ & $2.5\times10^4$ & $>10^6$ \\
full model ($V=1$ for $\rho$) & $4.5\times10^3$ & $6.2\times10^3$ & $1.0\times10^3$ & $1.6\times10^4$ & $2.0\times10^4$ & $2.6\times10^3$ \\
\bottomrule
\end{tabular}}
\end{table}

\paragraph{The intensity.} Each step up the ladder lowers the error at every $n$ (Fig.~\ref{fig:ladder}a). From the
band upwards, the KL falls close to $1/n$. Over $10^3$ to $10^6$ photons its slope is $-1.0$ for the band and the
two-slit shape and $-1.1$ for the full model. Between $10^5$ and $10^6$ the cross-validated kernels flatten to about
$-0.8$, because their selected cut-off and fill move on a grid. The KDE's error falls as $n^{-0.78}$, close to the
$n^{-4/5}$ expected for a kernel of fixed shape, whose bias in $h^2$ trades against a variance in $1/(nh)$. At $10^5$
photons the KL is $2.5\times10^{-3}$ for the KDE, $4.6\times10^{-4}$ with the band, $1.9\times10^{-4}$ with the
two-slit shape, $1.6\times10^{-4}$ with the known geometry and $4\times10^{-5}$ with the full model
(Table~\ref{tab:synth}). Put in photons (Table~\ref{tab:laddern}): reaching a KL of $10^{-3}$ takes $2.8\times10^5$
photons with the KDE, seven times fewer with the band, and half as many again with the slit shape. The full model needs
only $4.5\times10^3$; the difference is the price of not knowing the profile of the slits.

\begin{figure*}[t]
\centering
\includegraphics[width=\linewidth]{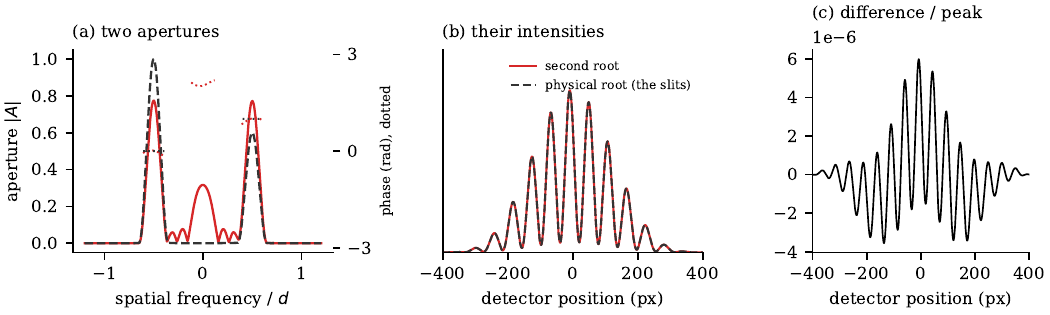}
\caption{Two apertures, one intensity. Complex BLML with the band at the true cut-off, fitted to the exact density of
the unequal slits of Fig.~\ref{fig:synthtruth}(b) (the limit of infinitely many photons). Red: the fit from its
default start. Dashed: the fit started from the true field. (a)~The two apertures, $|A|$ (solid) and phase (dotted).
One is the physical pair of slits, with amplitude ratio $0.6$ and a phase step of $1$\,rad; the other has nearly
equal slits with light between them. (b)~Their intensities on the detector. (c)~The difference between the two
intensities, a few millionths of the peak. The log-likelihoods of the two fits differ by $10^{-8}$ nats per particle.
The two-slit kernel rules out the second aperture.}
\label{fig:roots}
\end{figure*}

\paragraph{What the band gives.} The band alone is enough for the fringe contrast and for the relative phase of the
slits. Every estimator from the band upwards reaches a contrast error of $0.005$ at $2$--$5\times10^4$ photons and a
phase error of $0.01$\,rad at $1.4$--$2.1\times10^4$, and the full model does no better. The KDE's contrast is fifteen
times worse from $10^5$ photons on, because its bandwidth flattens the fringes.

The band does not reliably give the aperture, because it allows roots that have the right intensity but are not the
physical field (Sec.~\ref{sec:model}). For the equal slits, whose field is real, it finds the slits in about half the
data sets. For the unequal slits it never does: its root has a third peak between the slits and makes them almost
equally strong ($\rho\approx0.95$) and about three times too wide, at every $n$, and fitted to the exact density it
returns the same root. This is not the optimiser: started at the true field, the fit stays there, and the two roots
have the same likelihood to $10^{-8}$ nats per particle. Two different apertures, unequal slits with a phase step and
nearly equal slits with light between them, send their particles to the same places (Fig.~\ref{fig:roots}); no number
of particles separates them, and only knowing that the aperture is two slits excludes the second
\cite{crimmins1983uniqueness}.

\paragraph{The certificate.} Without a background mode, Theorem~\ref{thm:cert} applies. On the unequal slits both band
roots have $\lambda_{\max}/n=1.0000$, so both are certified global maxima on the data. For the two-slit kernel the
certificate flags a start that stopped at a local maximum ($2.05$, against $1.003$--$1.034$ for the starts that reach
the slits), so restarting until it is close to one is a practical rule (Supplemental Material, Sec.~S7).

\paragraph{What the slit shape gives.} The two-slit kernel removes the spurious roots. With its shape and fill chosen
by cross-validation, the separation is within $0.1\%$ from about $10^4$ photons on and the width within $5\%$ from
$5\times10^4$; cross-validation picks the true fill in 17 of 18 data sets at $10^5$ photons. Knowing the exact
geometry saves only a little more, up to a factor of three for the width. At $10^6$ photons, five of the 18 data sets
choose a wider fill, which shows up in the width. The reason is that the true cut-off is not on the grid of candidate
cut-offs; the known-geometry fit, which uses it, does not have this problem. The full model, which also knows the
slit profile, gets the width within $5\%$ already at $10^3$ photons, the fewest we simulated.

\paragraph{What no number of photons gives.} The two slits have the same profile, so by Eq.~\eqref{eq:gamma} the
intensity fixes only the contrast $\gamma=2\rho V/(1+\rho^2)$, not $\rho$ and $V$ separately
(Fig.~\ref{fig:ladder}f,\,g). With the prior $V=1$, the amplitude ratio of the unequal slits is recovered: within
$0.01$ from $2\times10^3$ photons with the known geometry, and within $0.001$ at $10^5$ (Table~\ref{tab:laddern}).
With the prior $\rho=1$, the full model recovers the visibility of the partially coherent truth within $0.001$ at
$10^5$. Without such a prior the estimates do not converge. For the unequal slits, the three-mode model returns an
amplitude ratio of $0.66$--$0.87$ from $10^4$ photons on, and a coherent share $\hat V$ of $0.88$--$1.0$, close to
their contrast of $0.88$. For the partially coherent truth its $\hat V$ ranges from $0.6$ to $1.0$, with a median of
$0.62$--$0.72$ from $10^4$ to $10^6$ photons, and the error does not fall with $n$. The full model with both
parameters free stays wherever it is started, because it sits in a valley of constant likelihood.

\subsection{The same truth on a record-like window}
\label{sec:crop}

Real records keep only a few fringes. We therefore take the same photons as above and keep only those that fall in a
window of seven fringes centred on the pattern, with both edges on dark fringes ($L=412$ pixels, the shaded band in
Fig.~\ref{fig:synthtruth}; $7\%$ of the signal lies outside it). We use $n=10^3$, $10^4$ and $10^5$, with three data
sets per truth and background level.

\begin{figure*}[t]
\centering
\includegraphics[width=\linewidth]{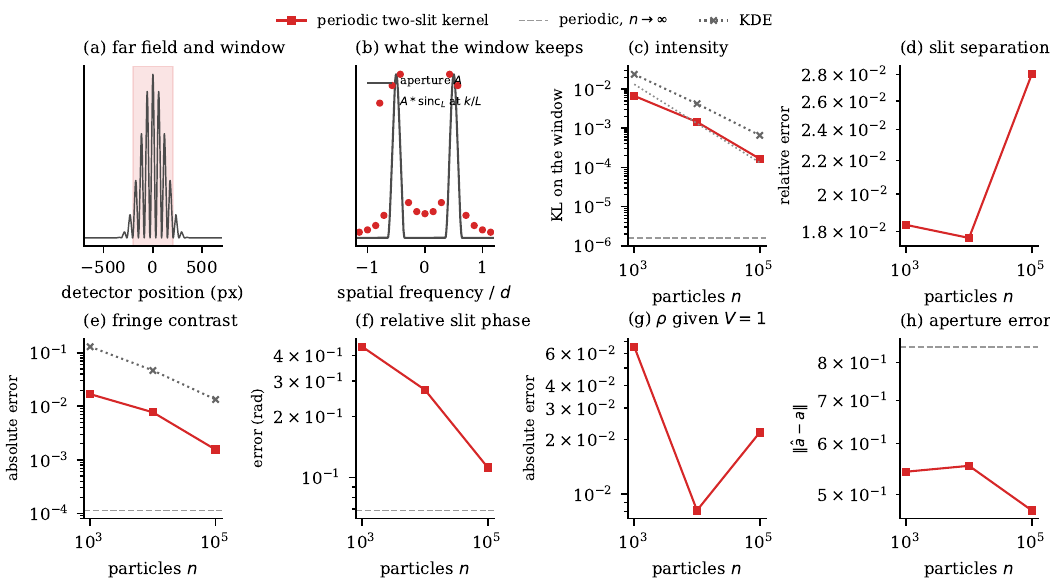}
\caption{The truth of Fig.~\ref{fig:synthtruth} on a window of seven fringes. (a)~The far field and the window.
(b)~What the window keeps: the aperture $A$ (equal slits) and the windowed truth $A*\mathrm{sinc}_L$ at the harmonics
$k/L$, two per slit. The remaining panels show errors against $n$ (median over six data sets): (c)~KL divergence on the
window (grey dotted line: slope $1/n$); (d)~slit separation read from the aperture; (e)~fringe contrast; (f)~relative
phase of the slits (unequal slits), measured about the fitted envelope centre; (g)~amplitude ratio from a coherent
model; (h)~aperture error against $A*\mathrm{sinc}_L$, minimised over the global phase, the twin and a shift of the
harmonics. The grey dashed lines show the periodic kernel fitted to the exact windowed density (infinitely many
photons), at the band with the lowest KL. Values for each truth are in the Supplemental Material, Sec.~S7.}
\label{fig:crop}
\end{figure*}

Cutting the pattern changes the truth itself. The data now come from the windowed intensity, whose root is
$g\,\mathbb 1_L$, with spectrum $A*L\,\mathrm{sinc}(L\nu)$ (Sec.~\ref{sec:model}). Sampled at the harmonics $k/L$ of
the window, each slit covers only two of them (Fig.~\ref{fig:crop}b), so the width and profile of the slits are no
longer in the data. We score every estimate against this windowed truth.

On the window the estimator is the periodic two-slit kernel, which keeps the harmonics
$k_{\rm lo}\le|k|\le k_{\rm hi}$ of the window and is normalised on it. Both limits are chosen by cross-validation,
together with a flat background. For the partially coherent truth we also fit the periodic three-mode model. We compare
these with the KDE. The two-slit kernel on the line is meant for patterns recorded whole. It normalises $|g|^2$ over
the whole line while the data lie only in the window, so the more photons there are, the more the likelihood penalises
mass outside the window. At $10^5$ photons it selects a plain band in all 18 data sets (Supplemental Material,
Sec.~S7). The rule is simple: use the line kernel when the whole pattern is recorded, and the periodic kernel on a
window.

\begin{table*}[t]
\centering\small
\caption{Complex BLML compared with the other methods on the truths of Fig.~\ref{fig:synthtruth}, at $10^4$ and
$10^5$ photons, both on the uncropped line (Sec.~\ref{sec:dbench}) and on the window of seven fringes
(Sec.~\ref{sec:crop}). Each entry is a median over six data sets per truth, averaged geometrically over the truths a
column uses. KL is the divergence from the truth (on the window, on the window). Separation and width are relative
errors (equal and unequal slits). Phase and amplitude ratio (given $V=1$) are for the unequal slits. Contrast is an
absolute error (unequal slits and the partially coherent truth). The two-slit kernel chooses its shape and fill by
cross-validation; the full physical model knows the slit profile and is fitted with $V=1$. On the window the slit
width is not in the data, and the separation is read from the aperture. HIO and the Fraunhofer and Fresnel models are
compared on a related benchmark in the Supplemental Material, Sec.~S7.}
\label{tab:synth}
\begin{tabular}{llrrrrrr}
\toprule
& $n$ & KL & separation & width & phase (rad) & contrast & $\rho$ ($V=1$) \\
\midrule
\multicolumn{8}{l}{\emph{uncropped, 2048 pixels}} \\
KDE & $10^4$ & 1.3e-02 & -- & -- & -- & 0.11 & -- \\
 & $10^5$ & 2.5e-03 & -- & -- & -- & 0.052 & -- \\
complex BLML, band & $10^4$ & 4.1e-03 & 9.1\% & 157\% & 0.015 & 0.016 & 0.37 \\
 & $10^5$ & 4.6e-04 & 5.3\% & 73\% & 0.0021 & 0.0031 & 0.39 \\
complex BLML, two-slit kernel & $10^4$ & 2.7e-03 & 0.11\% & 9.6\% & 0.014 & 0.013 & 0.0093 \\
 & $10^5$ & 1.9e-04 & 0.027\% & 2.2\% & 0.0049 & 0.0029 & 0.0033 \\
full physical model & $10^4$ & 4.2e-04 & 0.045\% & 0.26\% & 0.014 & 0.011 & 0.0094 \\
 & $10^5$ & 4.0e-05 & 0.016\% & 0.21\% & 0.0016 & 0.0017 & 0.0032 \\
\addlinespace
\multicolumn{8}{l}{\emph{window of seven fringes, 412 pixels}} \\
KDE & $10^4$ & 4.2e-03 & -- & -- & -- & 0.058 & -- \\
 & $10^5$ & 6.6e-04 & -- & -- & -- & 0.015 & -- \\
complex BLML, periodic two-slit kernel & $10^4$ & 1.4e-03 & 1.8\% & lost & 0.27 & 0.012 & 0.0081 \\
 & $10^5$ & 1.7e-04 & 2.8\% & lost & 0.11 & 0.0012 & 0.022 \\
\bottomrule
\end{tabular}
\end{table*}

\paragraph{What survives.} The periodic kernel still estimates the intensity, with a KL two to five times below the
KDE's at every $n$ (Fig.~\ref{fig:crop}c). It also recovers the fringe contrast, within $0.001$--$0.003$ at $10^5$
photons, compared with $0.010$--$0.017$ for the KDE. With a coherent model, the amplitude ratio of the unequal slits is
within $0.02$.

On the partially coherent truth, the three-mode model predicts no better than the single coherent mode, as
Theorem~\ref{thm:realroot} implies, but it keeps the slit separation where the single mode puts field between the
slits; its coherent share is again the contrast, not $V$ (Supplemental Material, Sec.~S7). What survives the window is
what lives in the fringes: the contrast, and with a prior, $\rho$ or $V$.

\paragraph{What is lost.} The aperture is not recovered (Table~\ref{tab:synth}). Against the window's own truth
$A*\mathrm{sinc}_L$, the aperture error of the periodic kernel is $0.44$--$0.69$ at every $n$. Fitting the exact
windowed density does not help: the band with the lowest KL is then the widest one tried, and its root is even further
from $A*\mathrm{sinc}_L$ ($0.63$--$0.96$). The windowed field is not periodic, so the periodic kernel can only
approximate it, and the intensity no longer singles out one root. The quantities read from the aperture suffer with
it. The relative phase of the slits is determined only to $0.1$\,rad at $10^5$ photons, twenty times worse than on the
uncropped line, and the fit to the exact density keeps an error of $0.06$--$0.1$\,rad. The separation of the slits in
the fitted aperture is off by $1$--$8\%$, compared with $0.03\%$ uncropped, although the fringe period still fixes
$d$.

A record cut to a few fringes therefore still determines the intensity, the contrast and, with a prior, the amplitude
ratio or the visibility. It does not determine the width or profile of the slits, and it determines their relative
phase only coarsely, however many particles are recorded. The missing information is in the tails that the window
removes. A record that keeps them, like the uncropped line, determines all of these quantities.

\section{A known aperture: the double slit}
\label{sec:dslit}

A coherent wave passes through two slits of width $a$ at separation $d$. In the far field its amplitude is the Fourier
transform of the aperture, so it is band-limited, with a band set by the extent of the aperture. Each detected
particle is a sample from the squared modulus of this amplitude. We ask what complex BLML recovers from the particles
alone, without a model of the slits.

\begin{figure*}[t]
\centering
\includegraphics[width=\linewidth]{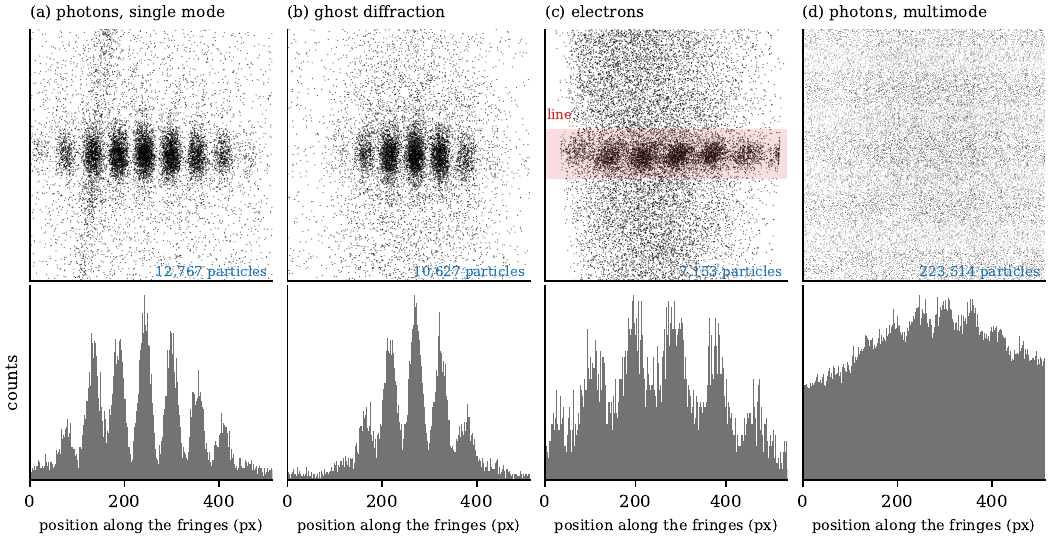}
\caption{The four double-slit records. Top: the detected particles on the detector, with the fringe axis horizontal.
For the photons we show the 256 rows that contain the beam; for the electrons, the detector around the 18-row strip
that holds the pattern (shaded). Bottom: the counts along the fringes, which are the data we fit.}
\label{fig:realdata}
\end{figure*}

\paragraph{Data and protocol.} Figure~\ref{fig:realdata} shows the four records. Aspden et al.\
\cite{aspden2016video} recorded single photons on a heralded, time-gated intensified camera behind two slits
($100~\mu$m wide, $500~\mu$m apart, $710$~nm): a coherent source (\PhN\ photons), a deliberately partially coherent,
multimode source (\MmN), and ghost diffraction (\GhN), in which the slits sit in the other arm of the photon pair and
the pattern appears only through heralding. Bach et al.\ \cite{bach2013controlled} built up a pattern one electron at a
time ($600$~eV, slits $62$~nm wide and $272$~nm apart; \ElN\ electrons). The pattern varies across the fringes only
through the beam profile, so we work on the line along the fringes, where by Theorem~\ref{thm:realroot} complex BLML is
correctly specified whether or not the beam is coherent; a background mode with $N$ harmonics across the line takes
the particles that are not signal. The records are cut to a few fringes, so following Sec.~\ref{sec:crop} the
estimator is the periodic two-slit kernel on a window trimmed to an odd number of half fringe periods. The particles
are split into training, validation and test sets (64/16/20\%), which for Poisson counts gives independent samples
\cite{neufeld2024data}; the harmonic range, $N\in\{0,1,2\}$ and the KDE's bandwidth are chosen by five-fold
cross-validation with the one-standard-error rule. The physical references are the two-slit Fraunhofer and Fresnel
models with free slit ratio, amplitude ratio, relative phase and visibility. Each method is scored by its
log-likelihood per particle on the test set, paired over test particles, over six random splits, and calibrated on
replicas of each record drawn from its Fresnel fit, for which the truth is known (details in the Supplemental Material,
Sec.~S6).

\begin{table*}[t]
\centering\small
\caption{All methods on the records' windows, for the paper's split. The first row gives the test log-likelihood per
particle of the periodic two-slit kernel; the other rows give each method's paired difference from it ($\pm$ SE). The
periodic kernel at the published slits uses the nearest harmonics (on the electrons and the multimode record these are
the selected ones). The periodic band keeps all harmonics up to the selected cut-off. The KDE, the adaptive KDE and the
two-slit models fit a flat background.}
\label{tab:dslit}
\resizebox{\linewidth}{!}{\begin{tabular}{lrrrr}
\toprule
 & photons, single mode & ghost diffraction & electrons & photons, multimode \\
window (pixels) / test particles & 392 / 2,449 & 384 / 2,084 & 459 / 1,362 & 393 / 36,437 \\
\midrule
periodic two-slit kernel, test LL & $-5.7736$ & $-5.6358$ & $-6.0557$ & $-5.9674$ \\
\addlinespace
\multicolumn{5}{l}{\emph{difference to the periodic two-slit kernel ($\pm$ SE)}} \\
periodic two-slit kernel, published slits & $-0.0006$ (0.0010) & $-0.0021$ (0.0015) & same harmonics & same harmonics \\
periodic band (no aperture) & $+0.0001$ (0.0001) & $-0.0014$ (0.0012) & $0.0000$ ($<$0.0001) & $0.0000$ ($<$0.0001) \\
KDE & $-0.0019$ (0.0013) & $-0.0012$ (0.0016) & $-0.0015$ (0.0016) & $-0.0001$ (0.0002) \\
adaptive KDE & $-0.0025$ (0.0013) & $-0.0004$ (0.0015) & $-0.0013$ (0.0016) & $-0.0002$ (0.0002) \\
Fresnel model & $-0.0078$ (0.0028) & $-0.0018$ (0.0012) & $-0.0001$ (0.0013) & $-0.0001$ ($<$0.0001) \\
Fraunhofer model & $-0.0099$ (0.0032) & $-0.0029$ (0.0016) & $-0.0008$ (0.0015) & $-0.0001$ (0.0001) \\
histogram & $-0.0172$ (0.0041) & $-0.0253$ (0.0051) & $-0.0368$ (0.0079) & $-0.0008$ (0.0003) \\
\bottomrule
\end{tabular}}
\end{table*}

\paragraph{Prediction.} On every record the periodic two-slit kernel predicts held-out particles at least as well as
every other method (Table~\ref{tab:dslit}). On the paper's split it is ahead of the KDE and the adaptive KDE by up to
$0.0025$ nats per particle ($0.0001$--$0.0002$ on the multimode record). It is ahead of the Fresnel and Fraunhofer
models by up to $0.010$ on the single-mode photons, about three standard errors, as is the KDE: the standard two-slit
models miss details of this record, its outer orders and envelope, and a free blur does not repair them. A
band-limited estimator with no model of the optics thus serves as a check of the optical model. The periodic kernel is
ahead of the histogram by $0.017$--$0.037$ ($0.0008$ on the multimode record). The periodic band and the periodic
kernel at the published slits come within $1.4$ standard errors of it: at these record sizes the aperture prior barely
changes the prediction.

A gap in test likelihood is a difference between two KL divergences from the truth, so it shrinks like $1/n$ even
when one estimator is much more accurate; the accuracy itself is better measured on replicas, where the truth is known.
There the periodic kernel's KL divergence from the truth is half that of the KDE for the single-mode photons and for
ghost diffraction, a sixth for the multimode record, and equal for the electrons. On the records themselves, averaged
over six splits, its lead over the KDE is within one replica standard deviation of the replicas' prediction, and within
$2.3$ for ghost diffraction (Supplemental Material, Table~S11). The KDE's bandwidth lowers the fringe contrast by
6--10\%; the band limit does not.

\paragraph{What the window determines.} The fringe contrast is recovered on every record (Supplemental Material,
Table~S11): within $0.014$ of the truth
on the replicas, and within $0.012$ of the Fresnel fit on the records, with a standard deviation over the six splits
of $0.002$--$0.016$. The slit separation is recovered on the
replicas of the photon, ghost and electron records, within 0.6\%. On the real photon and ghost records, however, it
comes out 7--8\% below the value given by the fringe period, because cross-validation keeps harmonics between the slits
that the replicas do not need (Supplemental Material, Fig.~S4). On the multimode record, whose contrast is only $0.05$, the
separation is 5--10\% low on the record and its replicas alike. The slit width is not determined on any record: on
these windows each slit spans fewer than two harmonics. The relative phase of the slits is recovered: measured about
the same envelope centre, it lies within $0.02$--$0.10$\,rad of the Fresnel fit's on every record (one split), against a replica
error of $0.04$--$0.21$\,rad (Supplemental Material, Fig.~S4). On a window this phase is tied to where the envelope is
centred, and on the single-mode photons, whose envelope the Fresnel fit misses, the two place it ten pixels apart.

\paragraph{Amplitudes and coherence.} By Eq.~\eqref{eq:gamma} the records determine the fringe contrast $\gamma$, not
the amplitude ratio and the visibility separately: a single coherent mode explains a contrast below one with unequal
slits, the Fresnel fit with a visibility below one (Supplemental Material, Fig.~S4). Which applies is a prior. A
heralded single-mode source \cite{aspden2016video} is spatially coherent, yet its fringes fall short of full contrast
($\gamma=0.75$), for example through spectral bandwidth or detector blur. If the slits are taken to be equal, the
three-mode periodic model reports the contrast as a visibility: $0.47$--$0.61$ on the electron record, consistent with
the $0.48$ measured by Bach et al.\ \cite{bach2013controlled}, though not a sharp test, since on the synthetic truth this
model read $0.04$--$0.14$ high; the coherent alternative, slits of amplitude ratio $0.26$, fits equally well. On the
multimode record the window cuts the pattern at half its peak, and the coherence is resolved only on replicas drawn
without the cut (Supplemental Material, Sec.~S6).

\paragraph{An uncropped record, and further checks.} Replicas drawn over a window widened by 1024 pixels on each side,
where nothing is cut, show what the cut costs: there the line two-slit kernel selects slits close to the published
ones, which the cut records do not allow, and on the multimode record the coherent share becomes resolvable
(Supplemental Material, Sec.~S6). Four further checks support the analysis (Supplemental Material, Sec.~S6). Complex
BLML selects the physical band ($1.06$--$1.30$ times it), identified from below. Its background mode recovers the
electrons' background measured in neighbouring rows ($\beta=0.32$ against $\ElBetaMeas$). Fitted to the first $n$
photons of the record as it builds up, it shows the two slits from $100$ photons and leads a KDE fitted to the same
photons by $0.02$ nats per photon at a few hundred. And successive detections are independent: the mutual information
between them is below $\IndPhBound$ nats on the single-mode photons.

\section{An unknown aperture: fluorescent beads}
\label{sec:beads}

\paragraph{Photon positions are samples from a band-limited density.} Let $P(\nu)$ be the pupil of a microscope
objective. It is zero outside $|\nu|\le\mathrm{NA}/\lambda$, and its phase carries the defocus and aberrations. A
point source produces the field $h=\mathcal F^{-1}[P]$ in the image plane. The intensity, $\mathrm{PSF}=|h|^2$, has the
autocorrelation of $P$ as its Fourier transform, which vanishes outside $|\nu|\le2\mathrm{NA}/\lambda$
\cite{goodman2005fourier}. A pixel integrates the intensity over its area, which multiplies the transform by
$\mathrm{sinc}(\nu_x)\mathrm{sinc}(\nu_y)$ without widening it, and out-of-focus light is smooth on the scale of the
PSF. The expected distribution of photons over the pixels is therefore a band-limited density, sampled at the pixel
centres.

\paragraph{A circular pupil with a known truth.}\label{sec:pupil} On simulated whole frames, complex BLML at the true
band estimates the intensity 3 to 25 times more accurately than the photon histogram, while the positive root of real
BLML is biased and is the worst estimator in every in-focus case (Supplemental Material, Sec.~S3). For recovering the
pupil phase from a single image with a known support, HIO phase retrieval is as accurate or better, also when complex
BLML uses the pupil's exact model, the periodic disk of harmonics (Sec.~\ref{sec:model}). The likelihood is the
reason: at $10^5$--$10^6$ photons, fits whose phase errors differ by up to $0.12$\,rad have likelihoods within a few
$10^{-5}$ nats per photon, and the fit started at the true pupil, no more likely than the others, is more accurate than
HIO at $10^6$ photons on three of four aberrations. One image determines the
finer pupil phase only weakly, and a prior on the phase would be needed (Supplemental Material, Sec.~S4). What complex
BLML adds is the likelihood itself: it chooses the band from held-out photons when the support is not known, and it
scores models on held-out data, as in the rest of this section.

\paragraph{Data.} We use the example bead data of the \texttt{localize-psf} package \cite{beaddata2023}: the plane of
best focus of a widefield $z$-stack (NA 1.3, 65\,nm pixels, sCMOS), converted to photons with a photon-transfer fit
\cite{janesick2007photon}, with the \NBeads\ beads that have no other local maximum within 16 pixels. Each pixel's
photons are split at random into training, validation and test counts, which for Poisson counts gives independent
images \cite{neufeld2024data} (Supplemental Material, Sec.~S5).

\begin{figure*}[t]
\centering
\includegraphics[width=0.92\linewidth]{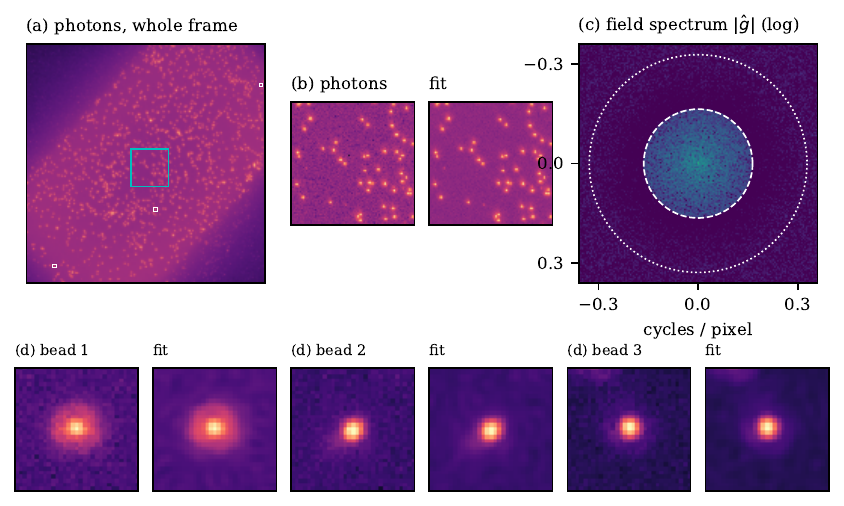}
\caption{Fluorescent beads. (a)~Training photons over the whole $2048\times2048$ frame (log scale; the cyan box marks
the region of panel b). (b)~That region: photons and the complex BLML fit of the whole frame at
$2\mathrm{NA}/\lambda$. (c)~The amplitude spectrum $|\hat g|$ of the fit (log scale). The spectrum of the field fills
the pupil disk $|\nu|\le\mathrm{NA}/\lambda$ (dashed) and nothing outside it, while the band of the intensity is twice
as wide (dotted). (d)~Three isolated beads, $31\times31$ pixels: photons and fit.}
\label{fig:beads}
\end{figure*}

\paragraph{The whole frame.} We fit the whole $2048\times2048$ frame (\FramePhotons\ million photons;
Fig.~\ref{fig:beads}) at the physical band $f_c=2\mathrm{NA}/\lambda$, without any selection and without a background
term: over the whole frame, the background is itself part of a band-limited density. We then crop each fit to the
\NBeads\ bead windows of $31\times31$ pixels, renormalise it there, and score it on each window's test photons. The
comparison is a Gaussian KDE fitted to the same window, with a uniform background floor and a bandwidth chosen on
validation data. Complex BLML beats the KDE on \FrameCxWins\ beads, ties on \FrameCxTies\ and loses on
\FrameCxLosses\ (median \FrameCxMed\ nats per photon). Real BLML at the same band loses on \FrameReLosses\ and wins on
none (median \FrameReMed). At the physical band, the positive root cannot follow the dark rings; this is the
obstruction of Theorem~\ref{thm:realroot}, in two dimensions. Adding the background as a second mode, the lowest
harmonic of the frame, changes nothing: it takes \FrameBgShare\% of the photons and moves the test score by
$\FrameBgDTest$ per photon. Over the whole frame, a smooth background is already band-limited.

\paragraph{The band.} When the frame is refitted at cut-offs from \FrameFcLo\ to \FrameFcHi\ times
$2\mathrm{NA}/\lambda$, the validation likelihood peaks at \FrameFcBest\ times the physical band, $\FrameFcGap$ nats per
photon above the fit at the band itself, and falls off on both sides. When the band is learned as the photons arrive
in batches, it settles at \FrameStreamFc\ times the physical band. The selection thus lands at or just below
$2\mathrm{NA}/\lambda$, where the transfer function of the microscope has fallen almost to zero. The certificate of
the frame fit is $\lambda_{\max}/n=\FrameCert$: close to one, but not a certificate of optimality.

\paragraph{Windows.} Fitting each $31\times31$ window on its own is harder, because the window cuts off the tails of
the bead and of its neighbours. With the background as a smooth second mode, which can absorb a neighbour's tail,
complex BLML beats the window KDE on \WinBgWins\ of \WinBgN\ beads and loses on \WinBgLosses\ (median
\WinBgMed\ nats per photon); with the periodic disk normalised on the window, the analogue of the periodic two-slit
kernel, it beats it on 27 and loses on 6 (median $+0.0016$). Both stay below the whole frame cropped to the same windows,
and with a flat background floor the windowed fits trail the KDE (Supplemental Material, Sec.~S5). Where the whole frame
is available, it should be fitted.

\section{Discussion}
\label{sec:discussion}

Complex BLML estimates the wavefunction behind detected particles from the one assumption the physics supplies, a
limited support in the conjugate domain, and accepts stronger priors where they are known. The positive square root,
the usual default, is band-limited only if the wave never changes sign, needs about twice the physical band, and
carries no phase. The simulations lead to three practical points, and the records bear them out.

First, each prior saves particles, and it is worth knowing how many. For two slits, the band gives the intensity with
seven times fewer photons than a kernel estimator, and it already gives the fringe contrast and the relative phase of
the slits. The shape of the aperture adds the slit geometry. Knowing the slit profile saves a further factor of about
four, and it is the only prior that gives the slit width well below $10^4$ photons. Table~\ref{tab:laddern} is the kind
of budget an experimenter can draw up before recording.

Second, some unknowns need a prior however many particles are recorded. For slits of the same profile the intensity
fixes only the fringe contrast, so an amplitude ratio requires a coherent source, and a visibility requires equal
slits. Without either prior, the coherent share of the three-mode model is not the visibility. On the electron record,
the prior of equal slits gives a visibility consistent with the one Bach et al.\ measured, but a coherent beam on
unequal slits fits as well. Defocus could separate the two, but defocus is itself poorly determined by one intensity on
the line; a second image at a known defocus would resolve it \cite{gonsalves1982phase}.

Third, cropping a record loses information that no estimator can recover. A window of a few fringes blurs the aperture
over a scale of $1/L$. The slit width is then no longer in the data, the relative phase is determined only coarsely,
and a model normalised over the whole line, which is the usual choice, is biased towards a plain band as $n$ grows. A
model normalised on the window avoids the bias but not the loss. On replicas of the records drawn without the cut, the
same fits recover the slits that the cut records do not. Where a camera records more than the fringes of interest, the
whole frame is worth keeping and publishing.

Some limits remain. A smooth background can trade against incoherent light, as on the multimode record, and against
the smoothness of the signal where the background is small; it is also assumed constant in time. The periodic kernel
does not contain a windowed field exactly, so even with unlimited particles its aperture is not that of the window; a
model normalised on the window that keeps the aperture of the line kernel would remove this floor. And the certificate
applies only to fits whose modes share one kernel, which excludes fits with a background mode; on real data it is a
diagnostic rather than a proof.

Two extensions lie well within what the estimator can already express, and we leave both to future work. The first
is a joint fit to particles recorded in two planes, for example near and far from the aperture or at two settings of
defocus. One $\psi$ then has to explain both records, linked by the propagator, so the likelihood tests the wave
equation itself, and the second plane removes most of the ambiguities of a single intensity, the twin included. The
second is the joint density of particle pairs. The ghost-diffraction record already depends on the correlation
between the two photons of a pair. If the positions of both photons were recorded, the same estimator applied to the
biphoton density $|\psi(x_1,x_2)|^2$, which is band-limited by the pump and the crystal, could estimate the
correlations that witness entanglement directly from the detected pairs.

\begin{acknowledgments}
We thank the authors of Refs.~\cite{aspden2016video,bach2013controlled,beaddata2023} for making their data public.
This research received no specific funding. R.A. is employed by Meta Platforms, Inc. A.A. is employed by Cohesity, Inc., and was previously employed by Rubrik, Inc. This work was carried out independently, outside the authors' employment; their employers had no role in it. The authors declare no other competing interests. 
\end{acknowledgments}

\section*{Data availability}
The double-slit photon records are those of Ref.~\cite{aspden2016video}, published by the University of Glasgow
(\url{https://doi.org/10.5525/gla.researchdata.281}); the electron record is that of Ref.~\cite{bach2013controlled}; and
the bead images are the example data of Ref.~\cite{beaddata2023}. The code is publicly available at
\url{https://github.com/rahuljhu/band-limited-ml} (MIT license), archived as release v1.0 at
\url{https://doi.org/10.5281/zenodo.24063291}; the processed data underlying the results are available
from the authors on request.

\end{document}